\documentclass[twocolumn]{article}

\usepackage{url}
\usepackage{times}
\usepackage{helvet}
\usepackage{courier}
\usepackage{amssymb}
\usepackage{amsmath}
\usepackage[caption=false,font=footnotesize]{subfig}
\usepackage{multirow}
\usepackage{hhline}
\usepackage{graphicx}
\usepackage[super]{nth}

\begin{document}
\title{Balancing Selection Pressures, Multiple Objectives, and \\ 
Neural Modularity to Coevolve Cooperative Agent Behavior}


\author{
        Alex C.\ Rollins \\
        Dept.\ of Mathematics and Computer Science\\
        Southwestern University\\
        Georgetown, TX 78626\\
        rollinsa@southwestern.edu
            \and
        Jacob Schrum\\
        Dept.\ of Mathematics and Computer Science\\
        Southwestern University\\
        Georgetown, TX 78626\\
        schrum2@southwestern.edu
}



%
%



\date{}

\maketitle

\begin{abstract}
Previous research using evolutionary computation in Multi-Agent Systems indicates that assigning fitness based on team vs.\ individual behavior has a strong impact on the ability of evolved teams of artificial agents to exhibit teamwork in challenging tasks. However, such research only made use of single-objective evolution. In contrast, when a multiobjective evolutionary algorithm is used, populations can be subject to individual-level objectives, team-level objectives, or combinations of the two. This paper explores the performance of cooperatively coevolved teams of agents controlled by artificial neural networks subject to these types of objectives. Specifically, predator agents are evolved to capture scripted prey agents in a torus-shaped grid world. Because of the tension between individual and team behaviors, multiple modes of behavior can be useful, and thus the effect of modular neural networks is also explored. Results demonstrate that fitness rewarding individual behavior is superior to fitness rewarding team behavior, despite being applied to a cooperative task. However, the use of networks with multiple modules allows predators to discover intelligent behavior, regardless of which type of objectives are used.
\end{abstract}

\section{Introduction}
\label{section:Introduction}

Evolutionary algorithms mimic real life evolution to develop solutions to difficult problems. Such algorithms allow teams of agents to develop complex, specialized behavior in evolutionary robotics, video games, and other agent-based simulations. 

Past research has studied the effects of selection pressures \cite{waibel:ieeetec09}, coevolution \cite{rawal:cig10,yong:ieeetamd10}, modular neural networks \cite{schrum:tciaig12,schrum:tciaig16}, and multiple objectives \cite{schrum:aiide08,vanhoorn:cig2009} in the evolution of complex agent behavior, but none of this research studies all at once. This paper explores how these concepts work in tandem. Various combinations of different types of selection (using multiple objectives) and different numbers of network output modules show how these components interact in the evolution of cooperative behavior. 


The concept of selection pressures in this research stems from prior research focusing on the rewarding of individual vs.\ team behavior \cite{waibel:ieeetec09}. For instance, if agents were part of a basketball team and their goal was to score as many points as possible, individual selection would reward each individual player based on the number of points that the individual scored, while team selection would reward each individual player based on the number of points that the entire team scored. So if the whole team performed poorly, but the individual player being assessed performed comparatively well, individual selection would grant that player higher fitness, while team selection would grant lower fitness. 

Multiple objectives are applied with the use of Pareto-based multiobjective optimization \cite{deb:tec02}. This framework allows both individual and team fitness functions to be used, thus going beyond the simple either/or comparison explored previously \cite{waibel:ieeetec09}. 
One goal of this research is to compare and contrast the effects of various types of selective pressures (individual, team, and both).

Studying the effects of modular networks is another goal.
Networks with multiple output modules can more easily generate multimodal behavior \cite{schrum:tciaig12, schrum:tciaig16}. Such modules can also make up for bad sensors \cite{schrum:tciaig16}, and similarly, this paper shows that these modules can also make up for bad fitness functions, allowing success for a team of agents even with less effective selection pressures.

This research utilizes cooperative coevolution, with separate and distinct sub-populations for each of the evolved agents. The sub-populations are evolved together, allowing them to develop specialized teamwork behaviors. This is in contrast to past research \cite{schrum:tciaig12}, which used multimodal networks with homogeneous teams. 

The next Section (\ref{section:Background}) describes related work and some background information that is relevant to this research. Section \ref{subsection:predPrey} describes the specifics of the predator/prey domain used. 
Section \ref{section:Evolution} describes the components of the evolutionary algorithm used. 
Section \ref{section:ExperimentalSetup} describes the experimental setup, while 
Section \ref{section:Results} provides and analyzes the results of the experiments. Section \ref{section:Discussion} discusses interesting discoveries as well as some ideas for future experiments. Finally, Section \ref{section:Conclusion} summarizes and concludes.

\section{Background}
\label{section:Background}

This research tests the effects of different selective pressures, multiple objectives, and modular networks on cooperatively coevolved agent behavior in a predator/prey domain. Each of these individual components has been studied in domains requiring intelligent agent behavior before, but they have not all been combined in one study. 




Some of the most relevant past research explored the connections between the evolution of team work, different types of team composition, and different types of selection pressures \cite{floreano:altruisticrobots2008,waibel:ieeetec09}. Specifically the effects of individual vs.\ team selection were explored using heterogeneous and homogeneous teams. One major finding was that heterogeneous teams performed poorly in cooperative tasks, but did better with individual selection. However, only single-objective evolution was used in these experiments, and teams were selected from a single population rather than separate isolated populations. Both of these distinctions have a meaningful impact in this paper.




Many domains have multiple objectives, so it is natural to apply Pareto-based evolutionary
methods \cite{deb:tec02}. However, it is still common to use single objectives instead,
even if this means creating a complicated fitness function that takes various different components
into account \cite{rawal:cig10,rajagopalan:cig11}. However, most team tasks have objectives
that measure individual performance as well as objectives that measure team performance,
which begs the question of which combination of such objectives will lead to the best results.
Although the merits of individual vs.\ team selection with multiple objectives 
have not been directly studied before, 
researchers are increasingly applying multiobjective approaches
to the kind of agent-based domain (mostly video games) that this paper focuses 
on~\cite{schrum:aiide08,vanhoorn:cig2009,schrum:tciaig12, schrum:tciaig16}. 




Some of this work has also focused on how to develop multimodal agent behavior using
both multiple objectives and modular neural network 
controllers \cite{schrum:tciaig12, schrum:tciaig16}.
There are many different concepts of modularity that 
are relevant to the evolution of neural 
networks \cite{clune:royal2013,kashtan:nasusa2005,verbancsics:gecco2011,schrum:tciaig16}. 
This research utilizes networks with explicit output modules that can be selected by an agent on each time step. This technique has been applied in situations similar to the predator/prey
domain used in this paper (Section~\ref{subsection:predPrey}): teams of homogeneous agents
were evolved to alternately attack/flee a scripted vulnerable/threatening 
opponent in a domain called Fight or Flight \cite{schrum:tciaig12},
as were skilled Ms.\ Pac-Man controllers that could both catch edible ghosts and flee
threatening ghosts \cite{schrum:tciaig16}. The need for multimodal behavior in these
domains came from the need to handle different dynamics when enemy agents switched roles
(attack vs.\ flee and catch vs.\ flee). In the predator/prey domain of this paper,
the need for multimodal behavior emerges from the tension between selfish 
and cooperative actions, which may be promoted in different ways by different types
of fitness functions.




Because this paper coevolves separate sub-populations that contribute
members to a single cooperating team, it is possible for the
available fitness functions to influence each sub-population differently.
Therefore, the teams in this paper are heterogeneous,
in contrast to those from the Fight or Flight study \cite{schrum:tciaig12}.
However, the previously mentioned research indicating that 
heterogeneous teams from a single population perform poorly
in cooperative tasks \cite{floreano:altruisticrobots2008,waibel:ieeetec09}
does not apply to these teams, because selecting individual team
members from specific sub-populations allows them to specialize in a way that
actually promotes cooperation \cite{gomez:ab97,potter:ec2000,Nitschke:2012,Nitschke:2013}.

Such cooperation is necessary for a team of evolved predator agents to capture fleeing prey agents in the domain described next.

\section{Predator/Prey Domain}
\label{subsection:predPrey}

Predator/prey scenarios have been used by researchers in many ways \cite{gomez:ab97,rawal:cig10,tan:icml93,Haynes1996,yong:ieeetamd10}. A survey by Stone and Veloso \cite{stone:MASsurvey} describes many variants of the simple predator/prey domain. Additionally, there are more complex domains that are essentially extensions of the predator/prey dynamic, such as the previously mentioned Fight or Flight domain \cite{schrum:tciaig12} and Ms.\ Pac-Man \cite{schrum:tciaig16}. 
However, even in a basic predator/prey domain, agents must exhibit intelligent cooperative behavior to succeed. 

The predator/prey domain of this paper is a torus-shaped grid world comprised of three predator agents attempting to catch two prey agents. The torus shape allows agents to wrap around from one edge of the world to the opposite edge. This design allows for infinite movement within a finite space. Consequently, even though the grid world has a fixed size the agents are always allowed to move in any direction. This makes catching the prey a difficult task for predator agents, as prey cannot be cornered or walled in. 

Distance in the grid world is measured by Manhattan Distance. 
The grid world itself consists of a 100 by 100 space square grid where agents each occupy a space; thus, the maximum horizontal or vertical distance from one agent to another is 50, and the maximum diagonal distance is 100. Agents can be located in the same space simultaneously, and if this is the case for a predator and a prey agent then the predator has caught the prey and the prey disappears. The domain also has a time limit of 1,000 time steps, where one time step is a single action for all agents (all actions happen simultaneously). The available actions are up, down, left, and right movements, as well as a null action (staying still). 

Each predator wants to maximize the number of prey it catches, but rewarding only this result will not be successful against competent prey. Therefore, use of other shaping objectives is common. The objectives in this paper are loosely based on Rawal et al.~\cite{rawal:cig10}, except that the complicated single objective from that work was split into simple components for multiobjective optimization.



Predators must work together in order to herd and capture prey. Selfishly chasing the prey generally leads to all agents going in circles around the torus. The success of the individual at least partially relies on the success of the team and the development of complex specialization. Predators develop jobs as valuable members of the team. Some common roles that emerge in successful teams are \emph{blocker}, \emph{herder}, and \emph{aggressor}. The blockers do not move very much but just align themselves at a distance with the side to side movement of the more aggressive predators so that they can force the prey to run toward the blocker. The herders work to keep the prey in front of the aggressors by running parallel to the prey's direction of movement, so that is does not slip by to one side. The job of the aggressor is to simply close the gap on the prey as quickly as it can. 


Though simple to describe, success in this domain is not trivial, which is why it has been widely studied by so many researchers. Therefore, sufficiently sophisticated methods are needed to evolve agents worth studying for this domain. The evolutionary methods used in this paper are described next.




\section{Evolutionary Algorithm}
\label{section:Evolution}


Predator agents were evolved using Modular Multiobjective Neuro-Evolution of Augmenting Topologies (MM-NEAT \cite{schrum:tciaig16}), which combines the multiobjetive evolutionary algorithm Non-Dominated Sorting Genetic Algorithm-II (NSGA-II \cite{deb:tec02}) and standard NEAT \cite{stanley:ec02}. MM-NEAT also allows for the evolution of networks with multiple output modules. MM-NEAT has been extended in this paper to support cooperative coevolution of separate sub-populations.

\subsection{Multiobjective Evolution}
\label{subsection:nsgaII}
 
When evolving intelligent agents, researchers typically use a single objective, but the objective is often complex and consists of several components. 
It is simpler and generally more effective to specify multiple objectives. 
Pareto-based multiobjective optimization provides a principled way of using multiple objectives that can discover trade-offs between objectives that are not attainable by the
common alternative of using a weighted sum. Even in comparison with single objectives
that are not weighted sums \cite{rawal:cig10}, the Pareto-based approach makes objectives
easier to define, and can also help evolution avoid local optima \cite{knowles:emo01}.
This approach depends on the concepts of Pareto Dominance and Pareto Optimality:

\noindent {\bf Pareto Dominance:}
Assuming a maximization problem, vector $\vec{v}=(v_{1},\ldots,v_{n})$ dominates vector
 $\vec{u}=(u_{1},\ldots,u_{n})$ iff

1.\ $\forall i \in \{1, \ldots, n\}: v_{i} \geq u_{i}$, and

2.\ $\exists i \in \{1, \ldots, n\}: v_{i} > u_{i}$.

\noindent Each vector is a collection of objective scores that an agent received during evaluation. 

\noindent {\bf Pareto Optimality:}
A set of points $\mathcal{A}\subseteq\mathcal{F}$ is Pareto optimal iff it contains all points such that
$\forall \vec{x} \in \mathcal{A}$: $\neg\exists \vec{y} \in \mathcal{F}$ such that $\vec{y}$ dominates $\vec{x}$.
The points in $\mathcal{A}$ are non-dominated, and make up the non-dominated Pareto front of $\mathcal{F}$.

The above definitions indicate that one agent is better than (i.e. dominates) another agent if it is strictly better in at least one objective and no worse in the others. The best agents are not dominated by any other agents, and make up the Pareto front of the search space. The next best individuals are those that would be in a recalculated Pareto front if the actual Pareto front were removed first. Layers of Pareto fronts can be defined by successively removing the front and recalculating it for the remaining individuals. Solving a multiobjective optimization problem involves approximating the first Pareto front as well as possible.

The multiobjective optimization algorithm used in this work is Non-Dominated Sorting Genetic Algorithm-II (NSGA-II \cite{deb:tec02}) which uses ($\mu + \lambda$) elitist selection favoring individuals in higher Pareto fronts (i.e. closer to the true Pareto front) over those in lower fronts. In the ($\mu + \lambda$) paradigm, a parent population of size $\mu$ is evaluated, and then used to produce a child population of size $\lambda$. Selection is performed on the combined parent and child population to give rise to a new parent population of size $\mu$. NSGA-II typically uses $\mu = \lambda$. 

When performing selection based on which Pareto layer an individual occupies, a cutoff is often reached such that the layer under consideration holds more individuals than there are remaining slots in the next parent population. These slots are filled by selecting individuals from the current layer based on a metric called \emph{crowding distance}, which encourages the selection of individuals in less-explored areas of the trade-off surface between objectives. 

By combining the notions of non-dominance and crowding distance, a total ordering of the population is obtained: individuals in different layers are sorted based on the dominance criteria, and individuals in the same layer are sorted based on crowding distance. The resulting comparison operator for this total ordering is also used by NSGA-II: Child populations are derived from parent populations via binary tournament selection based on this comparison operator. 

Applying NSGA-II to a problem results in an approximation to the true Pareto front. This approximation set potentially contains multiple solutions, which must be analyzed in order to determine which solutions fulfill the needs of the user. In this paper, the primary objective of interest is the number of prey captured. However, NSGA-II is indifferent as to how solutions are represented. In this paper, NSGA-II was used to evolve artificial neural networks to control the predators. The process of evolving these networks is called neuroevolution.

\subsection{Neuroevolution}
\label{subsection:NEAT}

Neuroevolution is the use of evolutionary algorithms to evolve artificial neural networks \cite{floreano2008neuroevolution}.
The evolved networks can be used to control agents in sequential decision making tasks by feeding in sensory input on every time step and interpreting the output for each time step as an action. This approach has been useful in many domains \cite{vanhoorn:cig2009, schrum:tciaig16,waibel:ieeetec09,gomez:ab97,Verbancsics:gecco11,huizinga:gecco2016}

The specific algorithm used in this work is a variant of Neuro-Evolution of Augmenting Topologies (NEAT \cite{stanley:ec02}) known as Modular Multiobjective NEAT (MM-NEAT \cite{schrum:tciaig16}). MM-NEAT combines the selection mechanism of NSGA-II with the network representation of NEAT, and adds additional features discussed in Section \ref{subsection:MM}. NEAT evolves artificial neural networks with arbitrary topologies. The networks begin with empty hidden layers and fully connected inputs and outputs, then evolution adds hidden neurons and new (potentially recurrent) links gradually via mutation in a process known as \emph{complexification}. 
Mutations can also change the weights of existing links.

Furthermore, every new link and neuron introduced by mutation is given a unique innovation number to identify it. 
The genotype that encodes each neural network stores these innovations linearly in a consistent order across all members of the population.
NEAT can perform efficient topological crossover by aligning genotypes based on these innovation numbers. 


Standard NEAT has been used to solve many challenging problems, but the resulting networks only define single control policies. The next section describes how MM-NEAT allows networks to have multiple policies, encouraging multimodal behavior.

\subsection{Modular Networks}
\label{subsection:MM}

Some of the networks in this paper can have multiple output modules. Each such module defines a different control policy. Arbitration between modules is discovered using special preference neurons that allow evolution to discover how to use the modules.


An output module is a collection of all output neurons needed to define the agent's behavior. These neurons are called \emph{policy neurons}. Each module also has one \emph{preference neuron}. Each module's preference neuron outputs the network's relative preference for using that module. Whenever inputs are presented to the network, the module whose preference neuron output is the highest is used to define the output of the network.

For example, the domain of this work requires 5 outputs to designate the behavior of an agent. Let us assume a given network has 2 modules. Then the network has 12 outputs: 5 policy neurons and 1 preference neuron for Module 1, and 5 policy neurons and 1 preference neuron for Module 2. Whenever the output of Preference Neuron 1 is higher than the output of Preference Neuron 2, the 5 policy neurons of Module 1 define the behavior of the agent. Otherwise, the policy neurons of Module 2 are used.

It is important for evolution to have the freedom to discover its own task division in the predator/prey domain due to the complex relationship between catching prey and closing in on the prey, and finding the right balance between the two. This means that an agent can have one specialized job as a part of the team at one time, and it can have completely different specialization at another time, and the system develops these behavioral responses situationally.

Support for preference neuron networks is one of the major innovations of MM-NEAT, but previous work with MM-NEAT only ever evolved a single population. MM-NEAT is extended in this work to support cooperative coevolution of multiple sub-populations as described next.

\subsection{Cooperative Coevolution}
\label{subsection:coevolution}

Coevolution is when the fitness of agents depends on other evolved agents. There are several models of coevolution, but the one used in this paper is cooperative coevolution with distinct sub-populations \cite{potter:ec2000, gomez:ab97}. Specifically, each team of agents is created by taking each team member from a separate sub-population, which makes it easier for specific team members to specialize into specific roles \cite{Campbell2011}. 




The coevolutionary process groups each genotype with random genotypes from the other populations to form different randomized teams. Having each genotype participate in several randomized teams addresses the structural credit-assignment problem \cite{agogino:aamas04} and ensures a more reliable evaluation of each individual. The structural credit-assignment problem arises when the success of a team could be the result of improved individual behavior or improved team behavior, and it is difficult to ascertain the source of the success. Consequently, it is uncertain how to best reward any particular individual vs.\ the entire team for the outcome. Having all individuals participate in multiple random teams means that individual's performance can be assessed more accurately. So, a bad agent is less likely to profit from getting lucky by being randomly placed with a good team and achieving good scores since this is unlikely to happen repeatedly.

Noisy evaluations are also relevant to this research, partly because the starting locations of agents are randomized. It is therefore even more important that each genotype be evaluated multiple times to assure reliable results.
    


\section{Experimental Setup}


Predators are evolved against scripted prey agents using several combinations of individual and team fitness functions, and numbers of output modules. The input sensors for the predators and most parameter settings remain constant across experimental runs. These details are discussed below.

\label{section:ExperimentalSetup}


\subsection{Agent Behavior}
\label{subsection:agent behavior}

Predators were allowed to do nothing (as an action) in addition to the four movement actions (up, down, left, right), but prey were restricted to the four movement actions. Predators could better act as the blocker when they did not have to learn to jump back and forth between the same location and could instead just stand still. This design decision is in line with previous work~\cite{rawal:cig10}. 


Predators act in accordance with their controlling neural networks.
In contrast, prey controllers are hard-coded to flee the nearest predator. These controllers first find which predator is the closest in terms of Manhattan Distance, then they calculate which of the four actions would result in the prey being the farthest possible distance from the current closest predator. 
The controller breaks ties randomly (another source of evaluation noise). So, if there are predators who tie as the closest in distance, the chosen predator is randomized among the set of the tying closest predators. Additionally, if multiple available movement locations are equally distant from the closest predator, then the specific movement is randomly chosen.

The static controller for the prey was simple to implement, but difficult for predators to learn to capture. Therefore, effective fitness functions are needed in order for the predators to succeed.


\subsection{Fitness Functions}
\label{subsection:fitness}


The goal of the predator teams is to maximize the number of prey captured across all evaluations, which means consistently capturing both prey within the time limit of each evaluation.
However, because the scripted prey behavior is fairly challenging (Section \ref{subsection:agent behavior}), predator agents evolved using only an objective that takes captures into account will not have a fitness gradient to follow when they fail to catch any prey, which is likely in the early stages of evolution when neural network genotypes are both simple and random.
Therefore, predators need at least some reward for \emph{almost}
capturing a prey agent. Such shaping can be accomplished with a distance fitness function, based on minimizing the final Manhattan Distances between predators and prey. Such concerns have influenced \emph{components} of a single-objective fitness functions used by others \cite{rawal:cig10,yong:ieeetamd10}.

This paper splits these types of objectives up into separate fitness functions, and also establishes individual and team versions of these objectives to evaluate a variety of different types of selection pressures. Predators can be evaluated by how many prey they personally catch, or the number the team catches. They can also be evaluated based on how close they are to prey when an evaluation ends, or by how close all predators are to the prey. Finally, multiobjective optimization makes it easy to combine all of these objectives as well.

Using only individual fitness functions applies different types of selective pressures than using only team fitness functions. Team fitness functions would seem more likely to promote team behavior, but perhaps some degree of selfish individual selection can actually lead to better overall group behavior. The concept of group selection is quite controversial in the realm of naturalistic evolution \cite{leigh:jeb2010}, but in a computer simulation it is straight-forward to test the effectiveness of such an approach without any concern as to its biological plausibility. Lastly, combining both types of selection seems as though it would be likely to provide the benefits of both. However, multiobjective optimization methods like NSGA-II are known to struggle when the number of objectives grows, since a higher-dimensional space is more likely to contain non-dominated points, making it difficult to make meaningful distinctions between candidate solutions. It is thus not obvious which scheme will be most effective in evolving effective behavior in the predator/prey domain.

The specific fitness functions used are defined as follows.
First recall that each evaluation contains three predators and two prey.
Then define $c_{i,j}$ to be $1$ if predator $i$
caught prey $j$ within the time limit, and $0$ otherwise.
Also, define $d_{i,j}$ to be the final
Manhattan Distance from predator $i$ to prey $j$, 
or $0$ if prey $j$ was ever caught by any
predator. These definitions are used to define
the fitness functions used in this paper.

The IndCatch objective defines how many
prey are caught by a particular predator.
Note that it is technically possible for two predators
to catch the same prey simultaneously.
For predator $i$,
\begin{equation}
\text{IndCatch}(i) = c_{i,0} + c_{i,1}  \label{eqn:IndCatch}
\end{equation}
A single objective, called TeamCatch, is also defined 
for the whole team to indicate how many prey were caught by
predators overall.
\begin{equation}
\text{TeamCatch} = \sum_{j=0}^{1} \max_{i \in \{0,1,2\}} c_{i,j}  \label{eqn:TeamCatch}
\end{equation}
The IndDist objective is defined for each combination of
one predator and one prey. If a predator cannot catch a prey,
it should at least decrease its distance from that prey by
the end of the evaluation. The best possible score in this objective
is zero, indicating that the prey was eaten.
Specifically, for predator $i$ with respect to prey $j$,
\begin{equation}
\text{IndDist}(i,j) = -d_{i,j} \label{eqn:IndDist}
\end{equation}
The team equivalent of this objective is TeamDist,
which is actually a set of objectives defined for each prey agent.
The objective measures the average distance of all predators
from one prey agent. Specifically, for prey $j$,
\begin{equation}
\text{TeamDist}(j) = - \frac{ \sum_{i=0}^{2} d_{i,j} }{3} \label{eqn:TeamDist}
\end{equation}






These individual fitness functions were combined in three ways summarized
in Table~\ref{tab:fitnesses}. The three specific groups of fitness
functions used focus either entirely on individual selection,
entirely on team selection, or on both:

\begin{enumerate}

\item Individual: There are three total fitness functions. For the population corresponding to predator $i$, the fitness functions used are $\text{IndCatch}(i)$,
$\text{IndDist}(i,0)$, and $\text{IndDist}(i,1)$.

\item Team: There are three total fitness functions. Each population uses the same fitness functions: $\text{TeamCatch}$, 
$\text{TeamDist}(0)$, and $\text{TeamDist}(1)$.


\item Both: There are six total fitness functions. For the population corresponding to predator $i$, the fitness functions used are $\text{IndCatch}(i)$,
$\text{IndDist}(i,0)$, $\text{IndDist}(i,1)$, $\text{TeamCatch}$, 
$\text{TeamDist}(0)$, and $\text{TeamDist}(1)$. Note that the final three of these
do not actually depend on $i$.


\end{enumerate}

\setlength\tabcolsep{1.5pt}
\begin{table}[tb]
  \caption{\small \label{tab:fitnesses}
        {\bf Objectives For Each Sub-population}}
  \begin{small}
  \begin{tabular}{| l || l | l | l | l |}
    \hline
      & IndCatch & IndDist & TeamCatch & TeamDist  \\ \hline \hline
Individual Selection  & {\tt 1} & {\tt 2} & {\tt 0} & {\tt 0} \\ \hline
Team Selection  & {\tt 0} & {\tt 0} & {\tt 1} & {\tt 2} \\ \hline
Both Selection  & {\tt 1} & {\tt 2} & {\tt 1} & {\tt 2} \\ \hline
  \end{tabular}
  \end{small} \\
{\small This table shows the number of fitness functions for each individual sub-population in each type of experiment. These numbers are the same for experiments where networks have either one or two modules. Ind stands for Individual Selection, and Team stands for Team Selection. Catch indicates the maximization of the number of prey caught. Dist indicates the minimization of distances between predators and prey (two distinct fitness functions of this type measure distances to the two distinct prey agents). The specific fitness functions used are defined in Equations \ref{eqn:IndCatch}, \ref{eqn:TeamCatch}, \ref{eqn:IndDist}, and \ref{eqn:TeamDist}.}
\end{table}

In summary, the number of prey caught is the primary metric of interest, but distance provides a fitness gradient when no prey are caught. Specifically, we care most about the number of prey caught by the team, but it is unclear whether rewarding team behavior, individual behavior, or both is the best way to achieve this goal. 
The use of modular networks can also have an influence on a team's ability to achieve this goal.

\subsection{Numbers of Modules}
\label{subsection:numModules}

\begin{figure}[t!]
  \centering 
\subfloat[One Module Network]{
  \label{fig:1M}
  \includegraphics[width=0.18\textwidth]{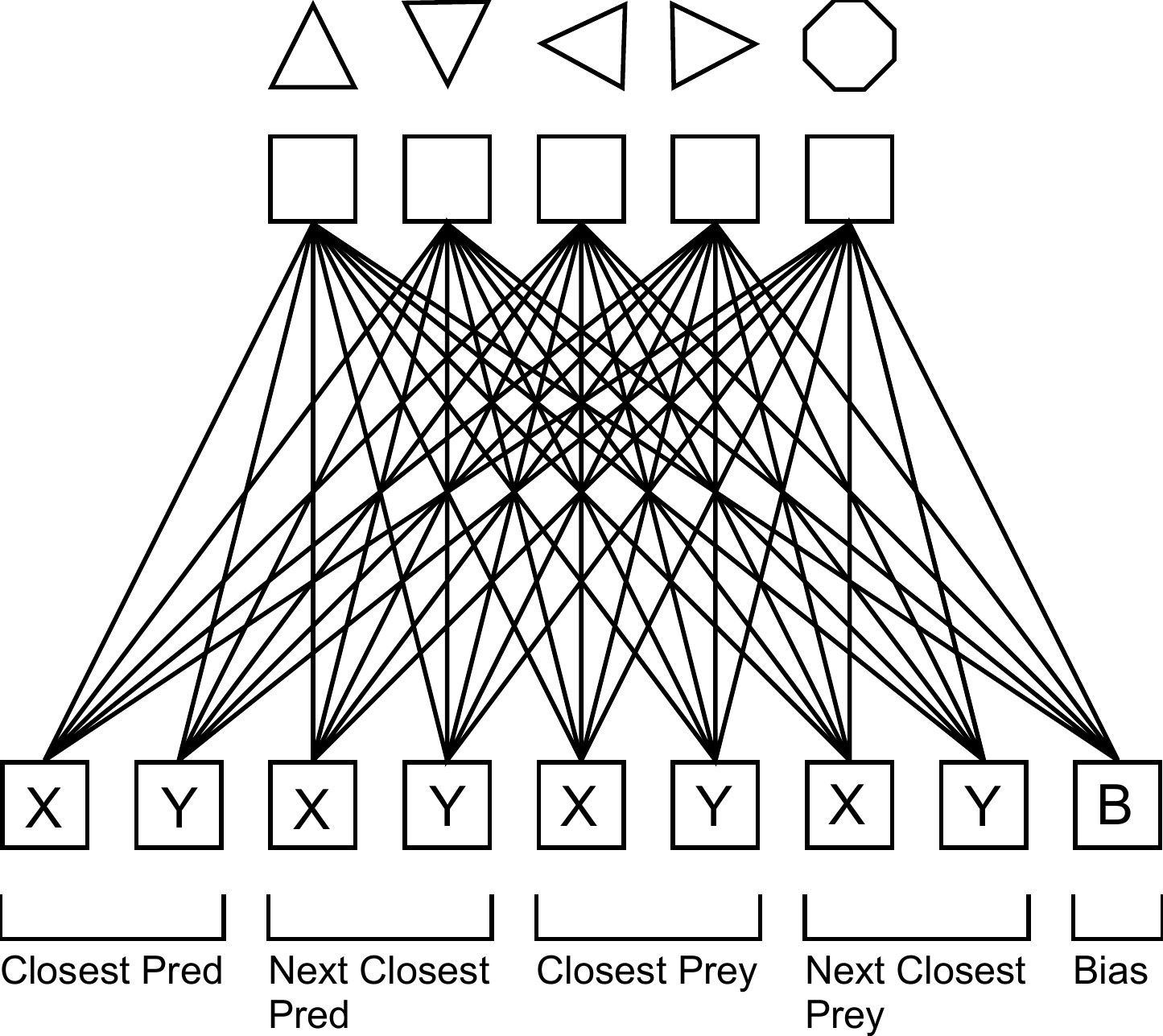}}
\subfloat[Two Module Network]{
  \label{fig:2M}
  \includegraphics[width=0.27\textwidth]{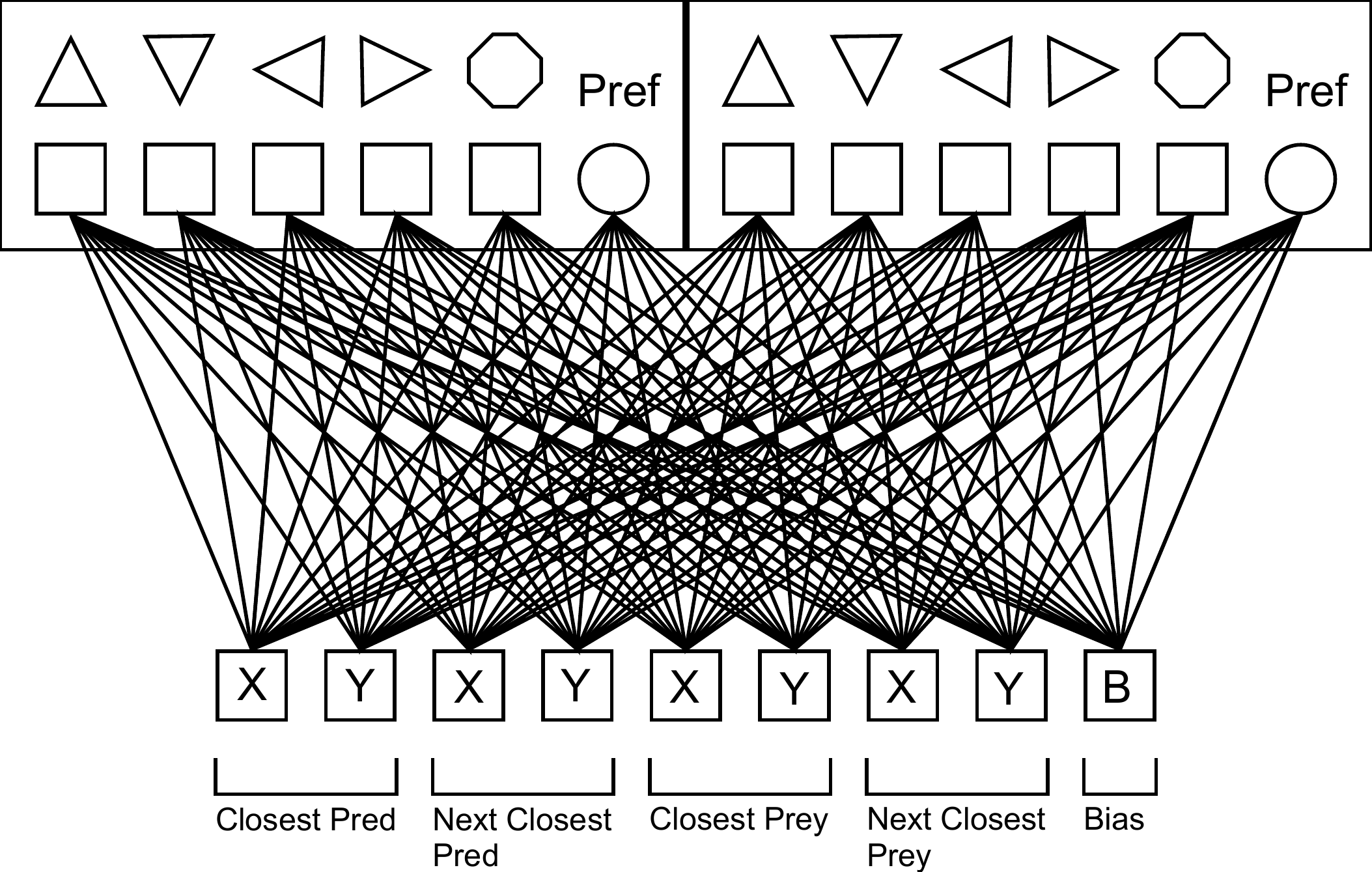}}
\caption{\small {\bf Starting Network Configurations.}
Both network configurations used by evolved predators are shown.
New populations start with no hidden neurons, but each output
is fully connected to all inputs.
\protect\subref{fig:1M} Networks with one module 
have outputs for moving up, down, left, and right,
as well as an output for staying still. Whenever inputs
are fed into the network, the agent it controls picks
the action with the highest output. The inputs are the
x/y offsets to each other agent in the grid world, followed
by a constant bias of 1.0. The agent inputs are grouped into
predators and prey, and sorted according to proximity in terms
of Manhattan Distance.
\protect\subref{fig:2M} Networks with two modules use
the same inputs, but have two distinct output modules.
Each module has all of the outputs possessed by the one
module network, as well as a preference neuron. For each
set of inputs, the two module network will pick the action
from the module whose preference neuron output is higher.
The additional module makes learning multimodal
behavior easier.}
\label{fig:networks}
\end{figure}

Each experimental setup in this paper utilizes either one or two modules, as shown in Figure~\ref{fig:networks}. Preliminary experiments were also conducted with more modules, but results indicated that any additional modules beyond two ended up being unnecessary, and were mostly ignored. 


Networks with one module (1M) are standard neural networks with a single behavior developed through evolution. Networks with two modules (2M) use preference neurons to switch between output modules, as described in Section \ref{subsection:MM}. 
Although these network types have different output configurations, they
both use the same input sensors, described next.

\subsection{Sensors}
\label{subsection:sensors}

Each predator's sensors are the normalized x and y distance offsets (in the range [-1.0, 1.0]) to each other agent. Since the predators are given the ability to sense teammates, the sensors include every agent (not just the prey), except the sensing agent. There is also a single constant bias input which always has a value of one. 
This means that there will be twice the number of sensor values as there are sensed agents, plus one for the bias, for a total of nine sensor values (x/y coordinates for each of two prey, and two predators besides the sensing predator). 
The sensor values become inputs into the network on each time step. 

The sensors are organized first by type (predator vs.\ prey) of agent being sensed, and secondly in ascending order of distance to each agent of that type. 
So within the predator and prey groups, the sensors begin with the closest agent in terms of overall Manhattan Distance, followed by the second closest, and so on. Also, when a prey is eaten, the distance to that prey from every other agent is set to the maximum distance (sensor value of 1.0), meaning that the other prey instantly becomes the priority. 



\subsection{Experimental Parameters}
\label{subsection:exp params} 
Combinations of the three different types of selection pressures discussed (Individual, Team, and Both) and the two different numbers of modules (1M or 2M) yield six experiments total. 
These experimental runs have the following labels: Individual1M, Individual2M, Team1M, Team2M, Both1M, Both2M.



The following settings were consistent across experiments.
Each experimental setup was run 30 times. 
There were three predators chosen from separate sub-populations, 
and two scripted prey. 
Every predator from each sub-population is evaluated in exactly ten randomly
chosen teams. Since each team results in a separate trial, this design
decision mitigates both the effects of noisy evaluations, and makes
fitness values more reliable in the face of the structural 
credit assignment problem.
Ten trials/teams were found to be enough to provide reliable evaluations for each genotype. 
Each experimental run lasts for $300$ generations. A population size of $\mu = \lambda = 200$ is used, so selection is performed across $400$ individuals. When offspring are produced, each network link has a 5\% chance of Gaussian perturbation. Additionally, each network has a 40\% chance of having a new random link added between existing neurons, and a 20\% chance of a new neuron being spliced along a randomly chosen link. Finally, topological network crossover has a 50\% chance of being applied when offspring are produced, with parents chosen via binary tournament selection. 
These settings lead to the results discussed next.


\section{Results}
\label{section:Results}

\begin{figure}[t!]
  \centering 
  \includegraphics[width=\columnwidth]{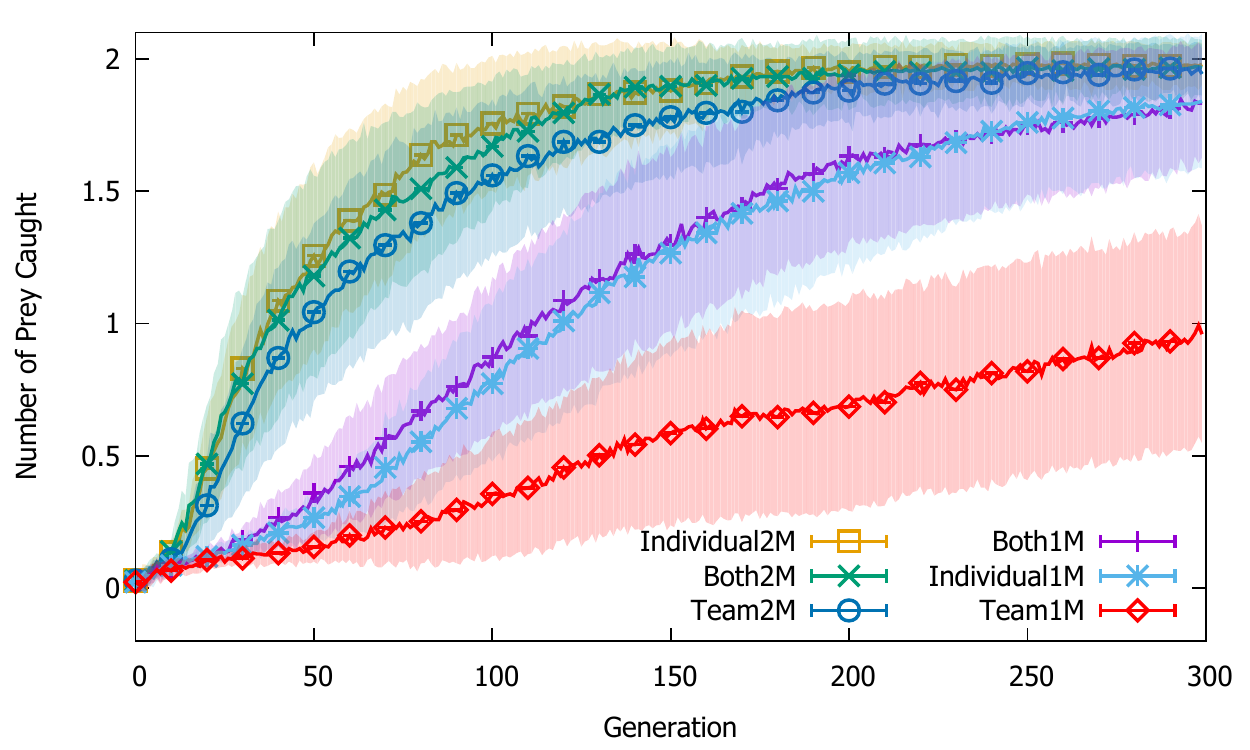}
\caption{\small {\bf Average Number of Prey Caught For Each Approach.} Average prey caught by champions across 30 runs of each method are plotted by generation with 95\% confidence intervals shown.
All 2M variants are superior to their 1M counterparts. Among 1M configurations, Both1M and Individual1M are superior to Team1M.}
\label{fig:Results}
\end{figure}

The results show that two modules is better than one module and that individual selection and combination setups are better than the purely team selection setup. Results during evolution are presented, followed by a discussion of the resulting behaviors. 


\subsection{Evolution}
\label{subsection:evolution}

Fitness plots of the average number of prey caught by the champion of each generation across 30 runs for each method are shown in Figure~\ref{fig:Results}. 
By comparing all 1M experiments to each other, results indicate that team selection is significantly inferior to other selection methods for this task ($p < 0.05$). 
Curiously enough, the performance of individual selection and the combination of individual and team selection were almost identical, and greatly superior to team selection. However, the final 1M performance is slightly short of perfect even for these two successful approaches.  

Every 2M variant is superior to its 1M counterpart. The starkest difference is for team selection, whose 1M variant is awful, but whose 2M variant is great (significantly better, $p < 0.05$). Every single predator team with two modules was able to reach a successful ceiling, capturing nearly all the prey on nearly every trial. The teams were even able to reach this point of optimization incredibly quickly, becoming almost completely leveled out by generation 150. Although the final performance levels of 2M runs are not significantly different from Individual1M and Both1M, they are better, and differences are significant ($p < 0.05$) for roughly the first 100 generations.



The reasons for the success of 2M methods can be understood
by analyzing the behaviors of evolved champions, discussed next.

\subsection{Behavior}
\label{subsection:behavior}

Behaviors of the champion agents are observed to see what kinds of behaviors predators develop to capture the prey. Additionally, for the 2M champions, movement paths were colored in a accordance with the modules being used in order to identify how behaviors were split up across modules.
Videos of representative behaviors can be seen at \url{southwestern.edu/~schrum2/SCOPE/predprey.html}.

The ability to easily switch between a more selfish Aggressor module and a more cooperative Support module is what allows 2M runs to succeed with all fitness combinations. In contrast, 1M champion teams tend to confine their specializations more to specific sub-populations and are unable to switch between different modes of behavior, which makes them less flexible, although Individual1M and Both1M teams eventually overcome this restriction because of effective selection pressures.

Some complexity is witnessed in the behavior of agents in even the worst of runs (i.e.\ Team1M).
In both good and bad runs,
all predators focus on the closest prey agent at the same time. 
One predator usually behaves as a \emph{blocker}. This behavior is essential in a torus world, because the prey needs to be surrounded and trapped in order to be captured. Successful teams included at least one agent which had this specialization. This predator typically moved in just a vertical or a horizontal line while the other predators performed the more complicated movements to force the prey to run towards/into the blocker.

Other predators herd the prey towards the blocker. This \emph{herding} behavior typically developed in each of the predator populations to at least some extent, though it was seldom the primary specialization. Rather, herding behavior was more of an auxiliary behavior that could be used as needed, at least within the more skilled populations, but it is still true that one predator will typically focus on herding behaviors more than the other two. This job includes chaotic, often side to side movements in relation to the targeted prey. Essentially, the herder moves parallel to the path of the prey's movement so that it will not escape in that direction. Simultaneously, this predator closes in on the prey whenever the movement does not sacrifice any of its herding positioning surrounding the prey. So, this second predator is a mix between the blocker and an aggressor, which is discussed next.

The role of \emph{aggressor} is the most prominent one across runs, likely because this behavior most obviously improves the distance fitness functions for an agent exhibiting this role. 
For some teams, this predator simply takes the movement action that will get it as close to the target prey as possible. In skilled populations, the aggressor is able to switch to a herding behavior when necessary, and will try to pick movements that help with the herding process (parallel to the prey). 

None of these specializations are strictly tied to one population. Predators can sometimes learn to take on different specializations at different times, as the need arises. For example, a predator could sometimes take on the roll of the blocker, and at other times be the aggressor. Such behavior is particularly prominent in agents with two preference modules.

The most prominent result from the fitness scores is the poor performance of Team1M. 
Therefore, it is not surprising that Team1M included more observable bad behaviors than the other setups. The most clearly visible example is that the predator team did not evolve specializations quite as strongly. These teams of predators did not
develop particularly focused blocking agents, as the closest thing to a blocker
was much more active in attempting to capture the prey, which made it harder for the other predators to herd and definitively surround the prey. Instead of confining the prey, these teams often let it slip through their grasp, which would then cause them to scatter before eventually homing in on the prey again.

In contrast, the Individual1M teams had clear blockers that made movements that corresponded to its teammates and stopped or slowed down a lot more often to allow its teammates to re-position and re-surround the prey. Since Team1M's closest thing to a blocker was moving so much, predators had to have more precise timing as they were closing in so that the prey wouldn't slip by them. This extra need for precise timing was avoided altogether by the Individual1M teams which ensured capture much more often with clear blocking behavior. 


The best teams consisted of agents with two preference modules, and the reasons for their success can be seen in their behavior, and how they use their modules. 
The predators regularly switch between the two modules in useful ways in every final champion 2M team. The two modules also appear to be dedicated to similar roles in each 2M run. 
One module is used when the predators are attempting to surround, 
herd, or block the prey, so this module will be referred to as the Support module. 
The other module is used when the predators are aggressively chasing the prey, so it will be called the Aggressive module. The Support module enacts long stretches of horizontal or vertical movements, without many changes in direction, since the predator is using the module to re-position. In contrast, the Aggressive module is used more often when the predators are closing in on the prey, so the predators' movements are as chaotic as those of the prey, with many changes in direction.

\section{Discussion and Future Work}
\label{section:Discussion}


The results clearly demonstrate the effectiveness of using multiple modules. This is due to the ability of agents to utilize two distinct behaviors whenever they are most helpful, specifically the effectiveness of having both selfish and cooperative behaviors at various times for this task. The predators were able to use one module (Support module) to re-position, surround, and be the blocker (supportive behavior) and one module (Aggressor module) to close in aggressively on the prey for the capture (selfish behavior).

Interestingly, the supportive behaviors may still lead directly
to individual fitness increases. Whether a predator is chasing
the prey, or blocking it, there will come a point when multiple
predators are equi-distant from the prey. At this point, due to
how prey agents randomly break ties when fleeing the nearest
predator, it is up to chance to determine whether the aggressive
or supportive predator captures the prey.



Results showed that one module team selection was quite ineffective for this task. Although, it is interesting that this ineffectiveness is overcome through the usage of multiple modules. Figure \ref{fig:Results} shows that even though team fitness functions are by far the least successful, they are still able to perform incredibly well when the networks are given two modules. It seems that with only one module available, team fitness functions will push agent behavior in a direction that makes the more complex blocking and herding behaviors more difficult to develop. However, given two modules, behaviors that provide only a marginal benefit early in evolution can be retained in a less used module until they have time to flourish in later generations.
The multimodal network capitalizes on only the positive effects from each of the selection pressures by not activating that particular behavior when it could perform better with the other one.

However, it is surprising that team fitness, which would seemingly focus on supportive behaviors and cooperation, would be less effective at developing blocking and herding behaviors.
The TeamDist fitness functions are perhaps to blame, because in assessing the whole team they are encouraging all predators to be close to the prey at the end of an evaluation. As a result, any blocking agents that were distant from the prey at the end of the evaluation would have lower fitness. Furthermore, predators that are close to the prey would be punished by teammates attempting to take on a blocking role, which would encourage even more aggressive behavior overall.
In contrast, the IndDist functions at least allow aggressors to not be punished by the behavior of blockers. Furthermore, since a population that tends toward blocking early in evolution will only be competing within its own niche, such a population would be pushed less strongly toward aggressive behaviors. There would still be a slight pressure toward such behaviors because of the IndDist functions, but since blocking behavior does eventually lead to more prey being caught, and therefore optimal IndDist values, a lessening of this selection pressure could provide enough generations for effective cooperation to emerge before blocking behavior is replaced with aggressive behavior.

Modular networks have already been frequently studied in conjunction with multiple objectives \cite{schrum:tciaig12,schrum:tciaig16}, but it would be interesting to see more work done combining these attributes with coevolution and various selection pressures, as is done in this paper. In particular, it would be interesting to see if the results regarding team selection are strongly tied to the specific fitness functions used in this paper, or are more general. 
Additionally, the effects of coevolution could be further studied by using competitive coevolution to also evolve the prey, in hopes of establishing an evolutionary arms race, as was done by Rawal et al.~\cite{rawal:cig10}. An alternative way of expanding this research would be to expand it to more complex domains, such as Ms.\ Pac-Man. Although MM-NEAT has already been used to generate successful Ms.\ Pac-Man behavior, it has not yet been used to generate behavior for the ghosts. Since the ghosts function as a team, cooperative coevolution could be applied. Furthermore, the role of different selection pressures would likely be important.

It would also be interesting to try a similar experiment with and without the sensing of teammates. This paper allowed the sensing of teammates, but past research \cite{yong:ieeetamd10} indicates that lacking such sensors can sometimes improve performance. According to this work, predators can coordinate effectively despite not sensing each other through \emph{stigmergy}, meaning that predators are able to take actions that complement the behaviors of fellow teammates by observing how the prey agents respond to all predators. That is, if a prey agent is approaching a predator despite being close, it stands to reason that it is being chased in that direction by a fellow predator.
We have conducted our own preliminary experiments that indicate the potential of stigmergy within our experimental setup as well.
Another means of changing the sensor configuration to adjust the challenge of the domain would be to organize the sensors with respect to particular agents rather than in terms of which agents are closest.

\section{Conclusion}
\label{section:Conclusion}

The predator/prey task is an interesting domain requiring teamwork and specialization. Results demonstrate that multimodal networks are extremely helpful. 
Additionally, this research indicates the superiority of individual selection in a teamwork oriented domain when coevolution across distinct sub-populations is used. This research also demonstrates the potential benefit of a mixture of individual and team selection. A combination is not only more flexible, but could also be the key to the full optimization of agent behavior. The usage of multiple objectives, multiple modules, coevolution, and the situationally appropriate selection pressures could be useful in more complex domains in the future.


\section*{Acknowledgments}

  This work is
  supported in part by the 
  Howard Hughes Medical Institute through the Undergraduate Science Education Initiative Program under Grant
  No.:~52007558.
  
  

\bibliographystyle{acm}
\bibliography{main} 

\begin{thebibliography}{10}

\bibitem{agogino:aamas04}
{\sc Agogino, A., and Tumer, K.}
\newblock Unifying temporal and structural credit assignment problems.
\newblock In {\em Autonomous Agents and Multiagent Systems\/} (July 2004).

\bibitem{Campbell2011}
{\sc Campbell, A., and Wu, A.~S.}
\newblock {Multi-agent Role Allocation: Issues, Approaches, and Multiple
  Perspectives}.
\newblock {\em Autonomous Agents and Multi-Agent Systems 22}, 2 (2011),
  317--355.

\bibitem{clune:royal2013}
{\sc Clune, J., Mouret, J.-B., and Lipson, H.}
\newblock {The Evolutionary Origins of Modularity}.
\newblock {\em Royal Society B 280}, 1755 (2013).

\bibitem{deb:tec02}
{\sc Deb, K., Pratap, A., Agarwal, S., and Meyarivan, T.}
\newblock {A} {F}ast and {E}litist {M}ultiobjective {G}enetic {A}lgorithm:
  {NSGA}-{II}.
\newblock {\em IEEE Transactions on Evolutionary Computation 6\/} (2002),
  182--197.

\bibitem{floreano2008neuroevolution}
{\sc Floreano, D., D{\"u}rr, P., and Mattiussi, C.}
\newblock Neuroevolution: {F}rom {A}rchitectures to {L}earning.
\newblock {\em Evolutionary Intelligence 1}, 1 (2008).

\bibitem{floreano:altruisticrobots2008}
{\sc Floreano, D., Mitri, S., Perez-Uribe, A., and Keller, L.}
\newblock Evolution of altruistic robots.
\newblock In {\em Computational Intelligence: Research Frontiers}. Springer
  Berlin Heidelberg, 2008, pp.~232--248.

\bibitem{gomez:ab97}
{\sc Gomez, F., and Miikkulainen, R.}
\newblock Incremental evolution of complex general behavior.
\newblock {\em Adaptive Behavior}, 5 (1997), 317--342.

\bibitem{Haynes1996}
{\sc Haynes, T., and Sen, S.}
\newblock Evolving behavioral strategies in predators and prey.
\newblock In {\em Adaption and Learning in Multi-Agent Systems}. Springer
  Berlin Heidelberg, 1996, pp.~113--126.

\bibitem{huizinga:gecco2016}
{\sc Huizinga, J., Mouret, J.-B., and Clune, J.}
\newblock {Does Aligning Phenotypic and Genotypic Modularity Improve the
  Evolution of Neural Networks?}
\newblock In {\em Genetic and Evolutionary Computation Conference\/} (2016),
  pp.~125--132.

\bibitem{kashtan:nasusa2005}
{\sc Kashtan, N., and Alon, U.}
\newblock {Spontaneous Evolution of Modularity and Network Motifs}.
\newblock {\em National Academy of Sciences 102}, 39 (2005).

\bibitem{knowles:emo01}
{\sc Knowles, J.~D., Watson, R.~A., and Corne, D.}
\newblock {R}educing {L}ocal {O}ptima in {S}ingle-{O}bjective {P}roblems by
  {M}ulti-objectivization.
\newblock In {\em International Conference on Evolutionary Multi-Criterion
  Optimization\/} (2001), Springer, pp.~269--283.

\bibitem{leigh:jeb2010}
{\sc Leigh~Jr, E.~G.}
\newblock The group selection controversy.
\newblock {\em Journal of Evolutionary Biology 23}, 1 (2010), 6--19.

\bibitem{Nitschke:2012}
{\sc Nitschke, G.~S., Eiben, A.~E., and Schut, M.~C.}
\newblock Evolving team behaviors with specialization.
\newblock {\em Genetic Programming and Evolvable Machines 13}, 4 (Dec. 2012),
  493--536.

\bibitem{Nitschke:2013}
{\sc Nitschke, G.~S., and Tolkamp, S.~M.}
\newblock Approaches to dynamic team sizes.
\newblock In {\em 2013 IEEE Workshop on Robotic Intelligence in Informationally
  Structured Space (RiiSS)\/} (April 2013), pp.~66--73.

\bibitem{potter:ec2000}
{\sc Potter, M.~A., and Jong, K. A.~D.}
\newblock Cooperative coevolution: An architecture for evolving coadapted
  subcomponents.
\newblock {\em Evolutionary Computation 8}, 1 (2000), 1--29.

\bibitem{rajagopalan:cig11}
{\sc Rajagopalan, P., Rawal, A., Miikkulainen, R., Wiseman, M.~A., and
  Holekamp, K.~E.}
\newblock The {R}ole of {R}eward {S}tructure, {C}oordination {M}echanism and
  {N}et {R}eturn in the {E}volution of {C}ooperation.
\newblock In {\em Conference on Computational Intelligence and Games\/} (2011),
  IEEE.

\bibitem{rawal:cig10}
{\sc Rawal, A., Rajagopalan, P., and Miikkulainen, R.}
\newblock Constructing competitive and cooperative agent behavior using
  coevolution.
\newblock In {\em Conference on Computational Intelligence and Games\/} (August
  2010), IEEE.

\bibitem{schrum:aiide08}
{\sc Schrum, J., and Miikkulainen, R.}
\newblock {C}onstructing {C}omplex {NPC} {B}ehavior via {M}ulti-{O}bjective
  {N}euroevolution.
\newblock In {\em Artificial Intelligence and Interactive Digital
  Entertainment\/} (2008), pp.~108--113.

\bibitem{schrum:tciaig12}
{\sc Schrum, J., and Miikkulainen, R.}
\newblock {Evolving Multimodal Networks for Multitask Games}.
\newblock {\em IEEE Transactions on Computational Intelligence and AI in Games
  4}, 2 (2012), 94--111.

\bibitem{schrum:tciaig16}
{\sc Schrum, J., and Miikkulainen, R.}
\newblock Discovering multimodal behavior in ms. pac-man through evolution of
  modular neural networks.
\newblock {\em IEEE Transactions on Computational Intelligence and AI in Games
  8}, 1 (2016), 67--81.

\bibitem{stanley:ec02}
{\sc Stanley, K.~O., and Miikkulainen, R.}
\newblock {E}volving {N}eural {N}etworks {T}hrough {A}ugmenting {T}opologies.
\newblock {\em Evolutionary Computation 10\/} (2002), 99--127.

\bibitem{stone:MASsurvey}
{\sc Stone, P., and Veloso, M.}
\newblock Multiagent systems: {A} survey from a machine learning perspective.
\newblock {\em Autonomous Robots 8}, 3 (July 2000), 345--383.

\bibitem{tan:icml93}
{\sc Tan, M.}
\newblock {M}ulti-{A}gent {R}einforcement {L}earning: {I}ndependent vs.
  {C}ooperative {A}gents.
\newblock In {\em International Conference on Machine Learning\/} (1993),
  pp.~330--337.

\bibitem{vanhoorn:cig2009}
{\sc van Hoorn, N., Togelius, J., and Schmidhuber, J.}
\newblock {H}ierarchical {C}ontroller {L}earning in a {F}irst-{P}erson
  {S}hooter.
\newblock In {\em Conference on Computational Intelligence and Games\/} (2009),
  IEEE, pp.~294--301.

\bibitem{verbancsics:gecco2011}
{\sc Verbancsics, P., and Stanley, K.~O.}
\newblock Constraining {C}onnectivity to {E}ncourage {M}odularity in
  {H}yper{NEAT}.
\newblock In {\em Genetic and Evolutionary Computation Conference\/} (2011),
  ACM, pp.~1483--1490.

\bibitem{Verbancsics:gecco11}
{\sc Verbancsics, P., and Stanley, K.~O.}
\newblock Constraining {C}onnectivity to {E}ncourage {M}odularity in
  {HyperNEAT}.
\newblock In {\em Genetic and Evolutionary Computation Conference\/} (2011).

\bibitem{waibel:ieeetec09}
{\sc Waibel, M., Keller, L., and Floreano, D.}
\newblock Genetic {T}eam {C}omposition and {L}evel of {S}election in the
  {E}volution of {M}ulti-{A}gent {S}ystems.
\newblock {\em {IEEE} {T}ransactions on {E}volutionary {C}omputation 13}, 3
  (2009), 648--660.

\bibitem{yong:ieeetamd10}
{\sc Yong, C.~H., and Miikkulainen, R.}
\newblock {C}oevolution of {R}ole-{B}ased {C}ooperation in {M}ulti-{A}gent
  {S}ystems.
\newblock {\em IEEE Transactions on Autonomous Mental Development 1\/} (2010),
  170--186.

\end{thebibliography}

\end{document}